\documentclass[letterpaper, 10 pt, conference]{ieeeconf}

\DeclareMathAlphabet{\altmathcal}{OMS}{cmsy}{m}{n}

\IEEEoverridecommandlockouts
\usepackage[colorlinks]{hyperref} 
\usepackage[utf8]{inputenc}
\usepackage{flushend} 
\usepackage{combelow}
\usepackage{amsmath}
\usepackage{tabularx}
\usepackage{amssymb}
\usepackage{mathptmx} 
\usepackage{subfigure}
\usepackage[export]{adjustbox}

\usepackage{color}
\usepackage{tikz}

\newcommand{\warning}[1]{\message{LaTeX Warning: #1}}
\newcommand{\ignore}[1]{\warning{Ignoring “#1”}}

\definecolor{CommentPMN}{rgb}{0.0,0.7,0.0}
\definecolor{CommentSS}{rgb}{0.0,0.4,0.6}
\definecolor{Blue}{rgb}{0.0,0.0,0.7}

\newcommand{\liningnums}[1]{#1}
\newcommand{\Threed}{\liningnums{3D}}
\newcommand{\Twod}{\liningnums{2D}}
\newcommand{\threed}{\Threed}
\newcommand{\twod}{\Twod}

\newcommand{\borg}{\textsc{bor\textsuperscript{\raisebox{-0.3ex}{2}}g}}

\newcommand{\gpu}{\textsc{gpu}}

\newcommand{\citeauthor}[2]{#2 \cite{#1}}
\newcommand{\citep}[1]{\cite{#1}}

\makeatletter
\newcommand*{\getlength}[1]{\strip@pt#1}
\makeatother

\usepackage[nolist]{acronym}

\begin{acronym}
  \acro{borg}[\borg{}]{Building Optimal Regularised Reconstructions with \gpu{}s}
  \acro{map}{\emph{maximum a posteriori}}
  \acro{vo}{Visual Odometry}
  \acro{imu}{Inertial Measurement Unit}
  \acro{slam}{Simultaneous Localisation and Mapping}
  \acro{sfm}[\textsc{s\textrm{f}m}]{Structure from Motion}
  \acro{ori}{Oxford Robotics Institute}
  \acro{glsl}{OpenGL Shading Language}
  \acro{mae}{mean absolute error}
  \acro{rmse}{root mean square error}
  \acro{cnn}{convolutional neural network}
  \acro{kitti-vo}{\textsc{kitti} visual odometry}
  \acro{jbf}{joint bilateral filter}
  \acro{tv}{Total Variation}
  \acro{tgv}{Total Generalised Variation}
  \acro{mlp}{multi-layer perceptron}
  \acro{dfn}{dynamic filter network}
  
\end{acronym}
\let\outsym\Delta

\def\idepthsym{d}

\newcommand{\pred}{\ensuremath{\outsym^*}}
\newcommand{\gtruth}{\ensuremath{\outsym}}

\newcommand{\predidepth}{\ensuremath{\idepthsym^*}}
\newcommand{\lqidepth}{\ensuremath{\idepthsym^{lq}}}
\newcommand{\hqidepth}{\ensuremath{\idepthsym^{hq}}}
\newcommand{\predidepthreproj}{\ensuremath{\widetilde{\idepthsym}^*}}
\newcommand{\weight}{\ensuremath{\mathrm{W}}}
\newcommand{\loss}[1]{\ensuremath{\altmathcal{L}^{#1}}}
\newcommand{\scaling}[1]{\ensuremath{\lambda^{#1}}}

\def\dataname{data}

\def\gradname{\ensuremath{\nabla}}
\def\gcname{gc}
\def\regname{reg}
\usepackage{pgfplotstable}
\usepackage{booktabs}

\pgfplotsset{compat=1.14}
\pgfplotstableset{
precision=2,
every head row/.style={before row=\toprule,after row=\midrule},
every last row/.style={after row=\bottomrule},
col sep=comma,
trim cells=true,
string type,
assign column name/.style={
    /pgfplots/table/column name={\textbf{#1}}
},
columns/imae/.style={ column name=i\textsc{mae}, sci, numeric type },
columns/irmse/.style={ column name=i\textsc{rmse}, sci, numeric type },
columns/mae/.style={ column name=\textsc{mae}, sci, numeric type },
columns/rmse/.style={ column name=\textsc{rmse}, sci, numeric type },
columns/thacc1.05/.style={ preproc/expr={100*##1} },
columns/thacc1.10/.style={ preproc/expr={100*##1} },
columns/thacc1.15/.style={ preproc/expr={100*##1} },
columns/thacc1.25/.style={ preproc/expr={100*##1} },
columns/thacc1.25^2/.style={ preproc/expr={100*##1} },
columns/thacc1.25^3/.style={ preproc/expr={100*##1} },
columns/delta1/.style={ preproc/expr={100*##1} },
columns/delta2/.style={ preproc/expr={100*##1} },
columns/delta3/.style={ preproc/expr={100*##1} },
thacc column/.style={%
    /pgfplots/table/columns/thacc#1/.append style={%
        column name=$\mathbf{\delta} < {#1}$, 
        numeric type, 
        fixed,
        fixed zerofill,
        dec sep align,
        postproc cell content/.append code={
            \ifnum1=\pgfplotstablepartno
                \pgfkeysalso{@cell content/.add={}{\%}}%
            \fi
        }
    } 
},
delta column/.style={%
    /pgfplots/table/columns/delta#1/.append style={%
        column name=$\mathbf{\delta} < {1.25^#1}$, 
        numeric type, 
        fixed,
        fixed zerofill,
        dec sep align,
        postproc cell content/.append code={
            \ifnum1=\pgfplotstablepartno
                \pgfkeysalso{@cell content/.add={}{\%}}%
            \fi
        }
    } 
},
delta column/.list={1,2,3},
thacc column/.list={1.05,1.10,1.15,1.25,1.25^2,1.25^3},
}

\title{Learning Geometrically Consistent Mesh Corrections}

\author{%
\cb{S}tefan Săftescu$^\dagger$ \and Paul Newman$^\dagger$
\thanks{$^\dagger$Oxford Robotics Institute, University of Oxford, United Kingdom; \newline 
  {\tt\small stefan,pnewman@robots.ox.ac.uk}}
}%

\begin{document}

\maketitle

\begin{abstract}

Building good \threed{} maps is a challenging and expensive task, which requires high-quality sensors and careful, time-consuming scanning.
We seek to reduce the cost of building good reconstructions by correcting views of existing low-quality ones in a post-hoc fashion using learnt priors over surfaces and appearance.
We train a \acl{cnn} model to predict the difference in inverse-depth from varying viewpoints of two meshes – one of low quality that we wish to correct, and one of high-quality that we use as a reference.

In contrast to  previous work, we pay attention to the problem of excessive smoothing in corrected meshes.
We address this with a suitable network architecture, and introduce a loss-weighting mechanism that emphasises edges in the prediction. 
Furthermore, smooth predictions result in geometrical inconsistencies.
To deal with this issue, we present a loss function which penalises re-projection differences that are not due to occlusions.
Our model reduces gross errors by 45.3\%–77.5\%, up to five times more than previous work.


\end{abstract}

\section{Introduction}\label{sec:intro}
\begin{figure}[t!]
\centering
\includegraphics[width=\columnwidth]{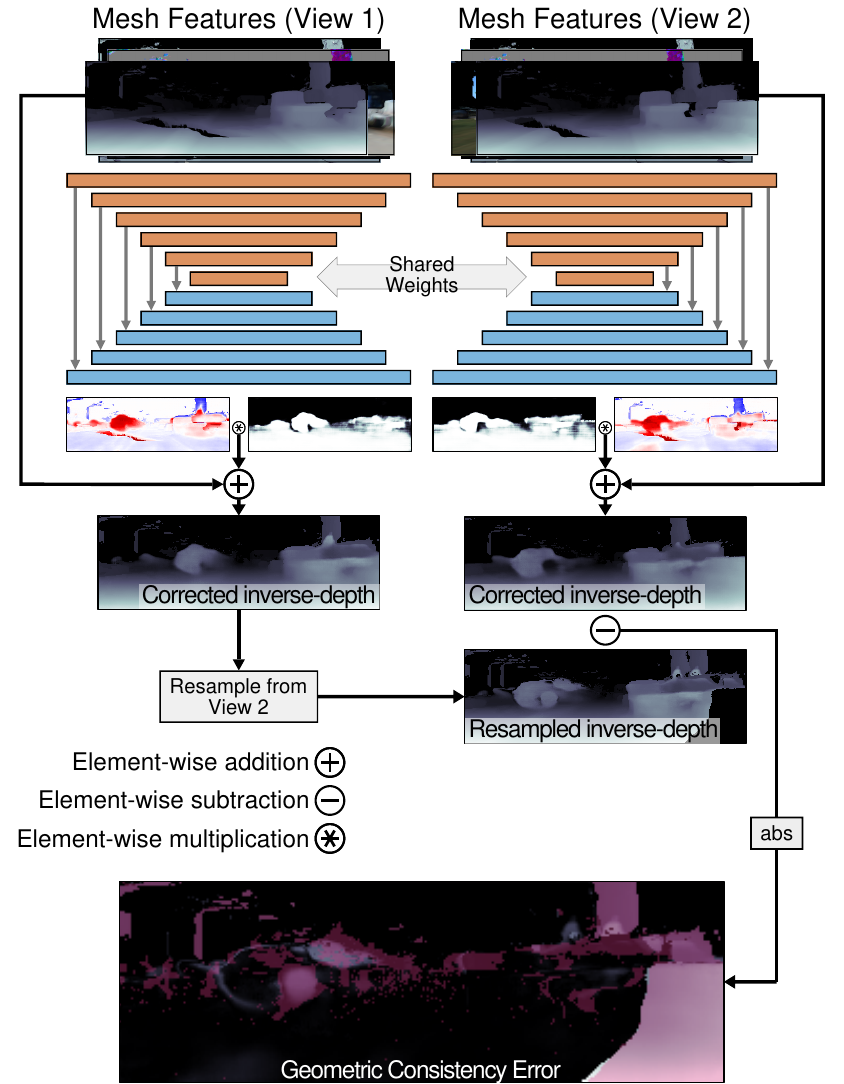}
\caption{
Illustration of geometric consistency for two views of the same scene a few meters apart.
The predictions of our model can be used to compute corrected depth maps for a set of views.
The corrected depth-maps should be consistent: they should be describing the same scene.
To enforce this, we densely reproject inverse-depth from View~1 to View~2, using the corrected depth of View~2.
The absolute difference (bottom) is the inconsistency of View~1 with respect to View~2.
This is minimised during the training of our model, which enforces geometrically consistent predictions.
The red overlay is the occlusion mask: the parts of the View~2 prediction that cannot be consistent with the View~1 prediction.
}
\label{fig:geometric_consistency}
\end{figure}

Dense \threed{} maps are a crucial component in many systems and better maps make robots easier to build and safer to operate.
Despite recent progress in hardware such as the wide availability of \textsc{gpu}s, and algorithms that scale with the amount of data and the available hardware, high-quality maps, especially at large scales, remain difficult to build cheaply, often requiring expensive sensors.

The main motivation of our work is to reduce the cost of building good dense \threed{} maps.
The reduction in cost can come from either:
a) cheaper but nosier sensors, such as stereo cameras; 
b) less data, and therefore less time spent densely scanning an area.
There are two ways in which we can produce better reconstructions with cheaper data. 
We can learn the kinds of errors a certain modality produces. 
For a stereo camera, for example, there will be missing data in areas without a lot of texture (walls, roads), and the ambiguity in depth is usually along the viewing rays.
In addition, we can learn priors for a target environment: cars usually have known shapes, roads and buildings do not have holes in them, surfaces tend to be vertical or horizontal in an urban environment, etc.

\begin{figure}[h!]
\centering
\subfigure[Colour Reconstruction]{%
\includegraphics[trim={0 1cm 0 2cm},clip,width=0.9\columnwidth]{%
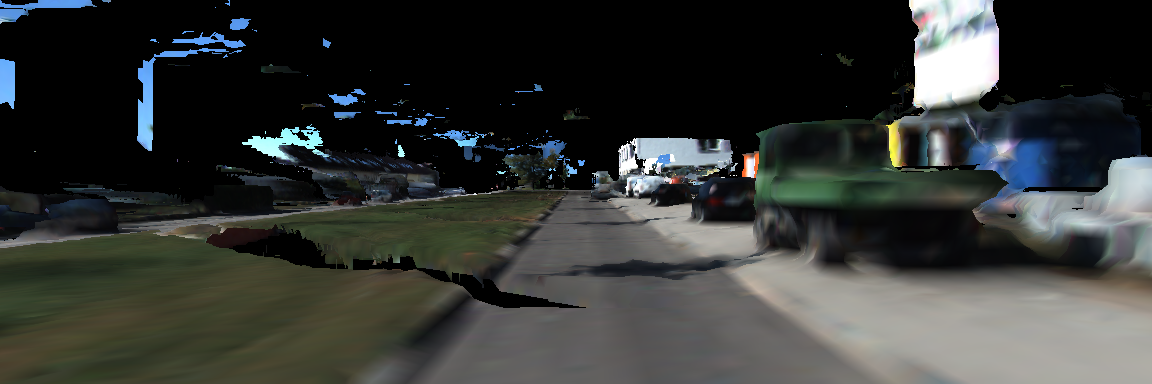%
}}
\subfigure[Inverse-depth]{%
\includegraphics[trim={0 1cm 0 2cm},clip,width=0.9\columnwidth]{%
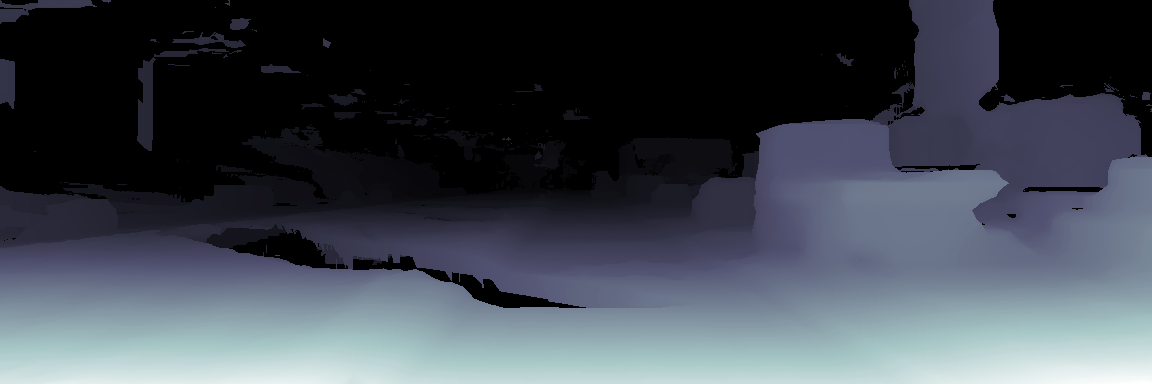%
}}
\subfigure[Triangle Surface Normal]{%
\includegraphics[trim={0 1cm  0 2cm},clip,width=0.9\columnwidth]{%
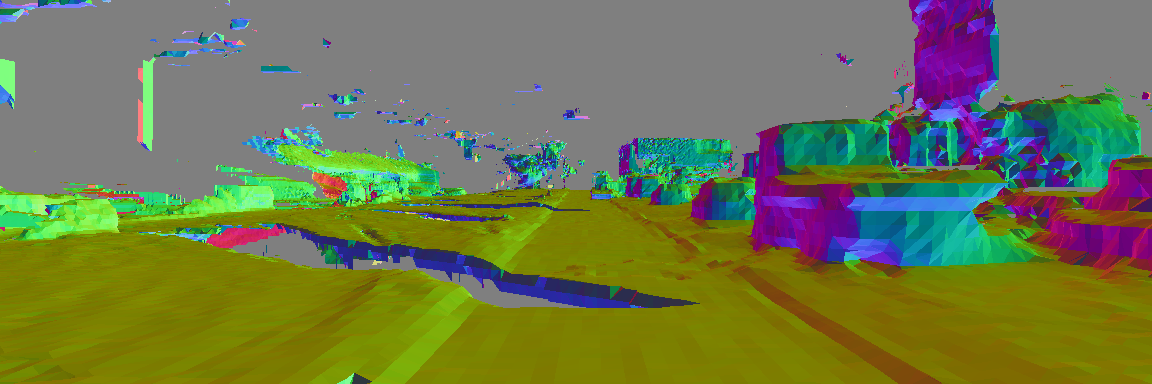%
}}
\subfigure[Triangle Area]{%
\includegraphics[trim={0 1cm  0 2cm},clip,width=0.9\columnwidth]{%
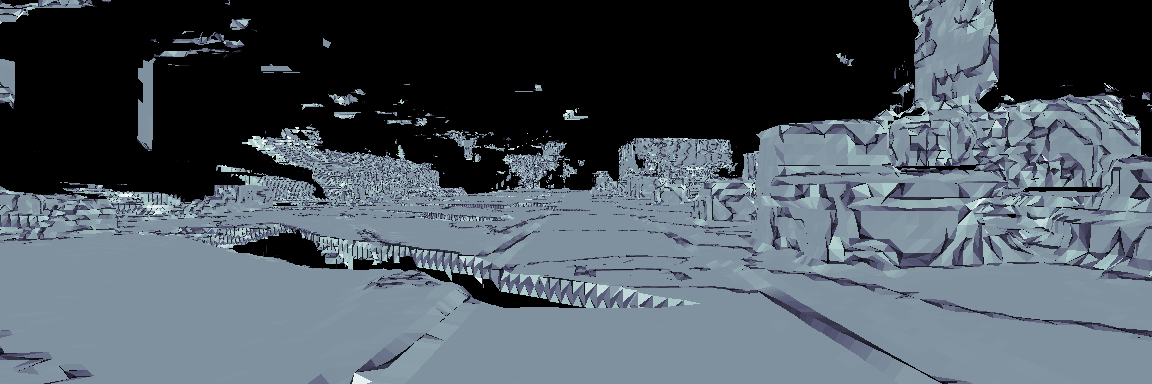%
}}
\subfigure[Triangle Edge Length Ratios]{%
\includegraphics[trim={0 1cm  0 2cm},clip,width=0.9\columnwidth]{%
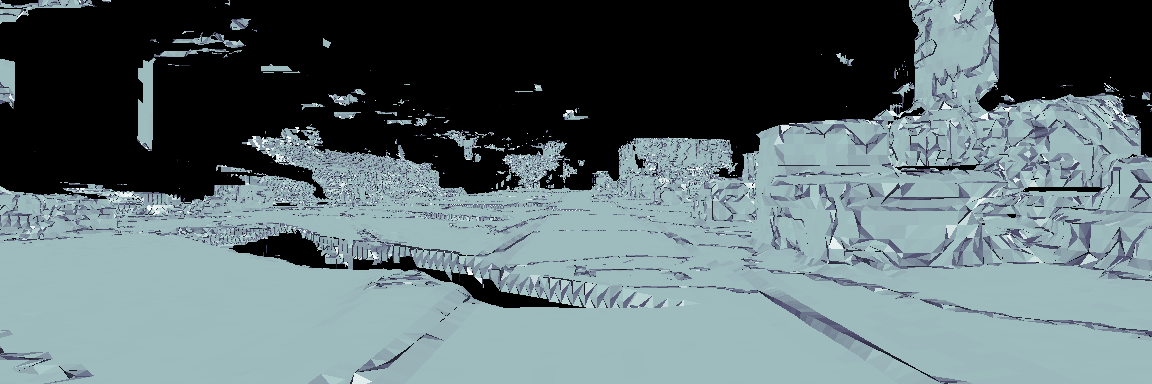%
}}
\subfigure[Surface to Camera Angle]{%
\includegraphics[trim={0 1cm  0 2cm},clip,width=0.9\columnwidth]{%
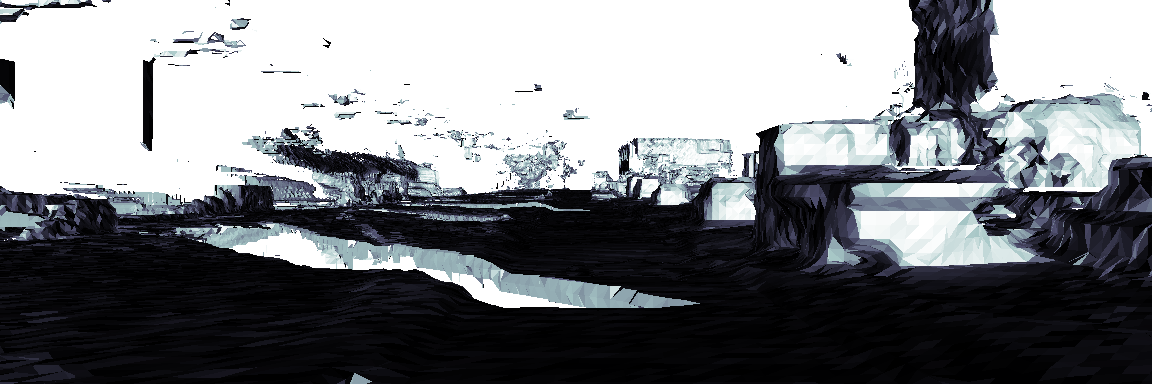%
}}
\caption{
Example mesh features.
We fly a camera through an existing mesh and at each location produce mesh features.
During training, inverse-depth images of a high-quality mesh are available, and our model learns a mapping from the mesh features pictured above to the high-quality inverse-depth.
We refer the reader to \citep{Tanner:2018:MeshedUp} for an analysis of the relative usefulness of mesh features.
}
\label{fig:mesh_features}
\end{figure}

We tackle the problem of correcting dense reconstructions with a \ac{cnn} that operates on rasterised views of a \threed{} mesh as a post-processing step, following a classical reconstruction pipeline.
To train this model, we start with two meshes, a low-quality one and reference high-quality one.
From each reconstruction, we render multiple types of images (such as inverse-depth, normals, etc.), referred to as \emph{mesh features} (shown in Figure~\ref{fig:mesh_features}), from multiple viewpoints. 
We then train the model on the mesh features to predict the difference in inverse-depth between the high-quality reconstruction and the low-quality one, thus enabling us to correct the low-quality mesh.

Previous work \citep{Tanner:2018:MeshedUp} has demonstrated the idea of correcting meshes post-hoc via \twod{} rasterised views.
We address its two main limitations.
Firstly, we deal with the issue of overly smooth predictions.
We propose some architecture and training changes: we add skip connections from the encoder to the decoder in our \ac{cnn}, which are known to help in localising edges in predictions; we propose a loss-weighting method that penalises incorrect predictions more the closer they are to an edge.
Secondly, predictions on nearby views are not always consistent, i.e. when applying the predicted corrections the geometry of the scene is not always the same.
To improve consistency, we employ a view synthesis based loss, and show that this also improves the performance of the network.
An illustration of this idea is shown in Figure~\ref{fig:geometric_consistency}.


Our contributions are as follows:

\paragraph*{Error correction}
We propose a  \ac{cnn} model that is able to correct \twod{} views of a dense \threed{} reconstruction. We propose a novel weighting mechanism to improve performance around edges in the prediction.
We evaluate it against existing work and show that we outperform it, especially when images are from multiple viewpoints.

\paragraph*{Geometric consistency}
We adapt the photometric consistency loss that features in related tasks such as depth-from-mono \citep{Godard:2017:UnsupervisedMonoDepth} to the task of correcting reconstructions. 
This is a novel use of the loss, which has thus far only been employed on \textsc{rgb} images. 
We leverage the existing reconstruction to compute occlusion masks, and thus exclude from the loss areas where geometric consistency is impossible. 
We show that this use of geometric consistency as an auxiliary loss further improves our model.

\begin{figure*}[ht!]
\centering
\subfigure[Front-left view.]{%
\includegraphics[trim={0 0  0 1cm},clip,width=\columnwidth]{%
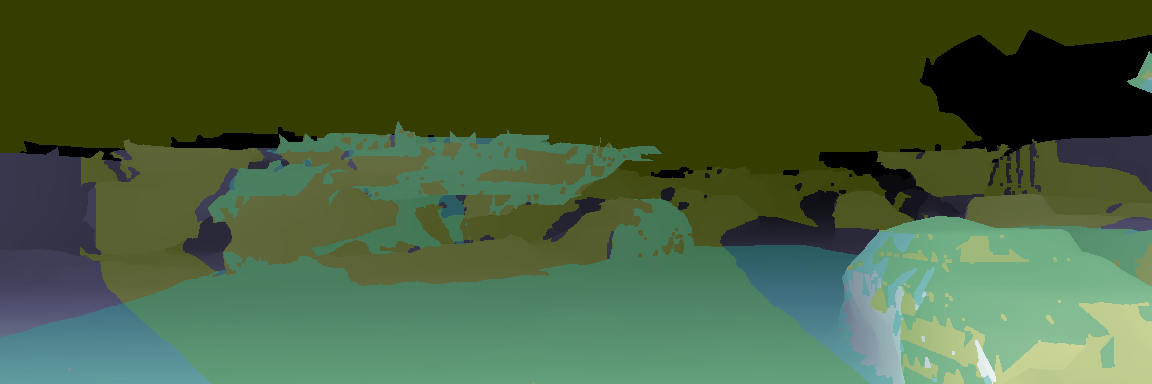%
}}\hfill%
\subfigure[Front-right view.]{%
\includegraphics[trim={0 0  0 1cm},clip,width=\columnwidth]{%
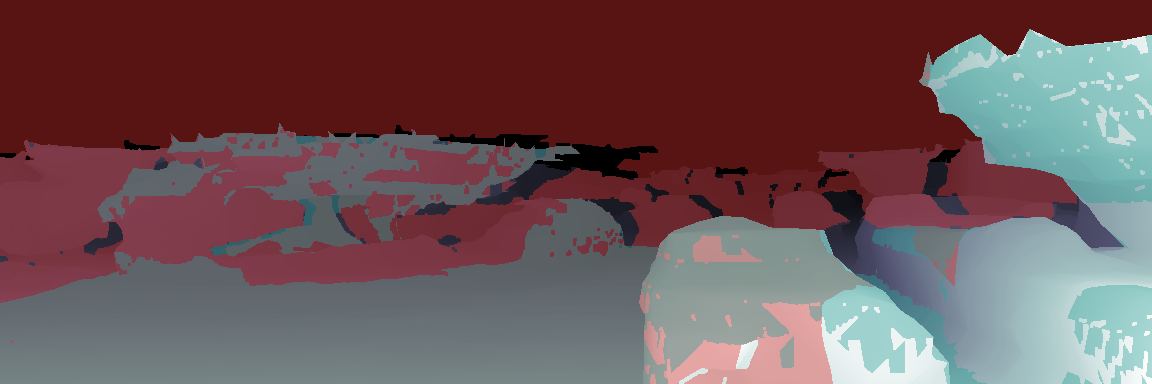%
}}\\
\subfigure[Back view.]{%
\includegraphics[trim={0 0  0 2cm},clip,width=\columnwidth]{%
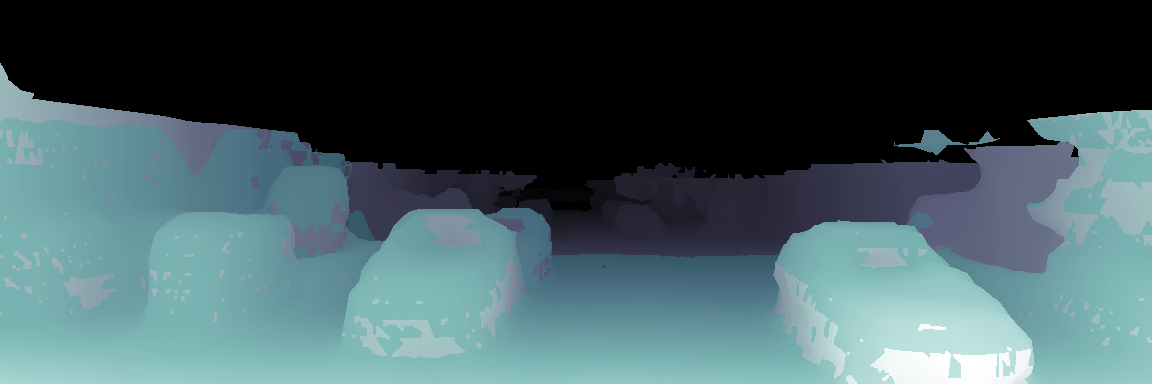%
}}\hfill%
\subfigure[Top view.]{%
\includegraphics[trim={0 0  0 2cm},clip,width=\columnwidth]{%
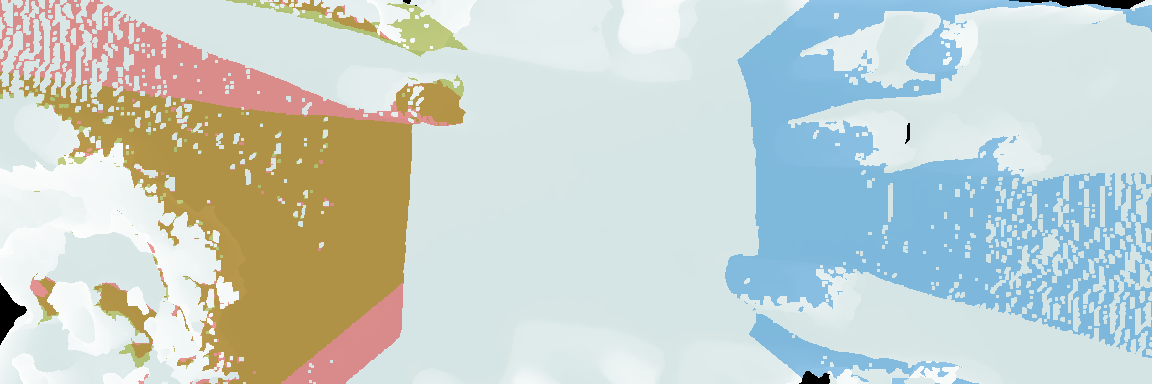%
}}
\caption{Visibility masks for views generated at a location. 
The image regions highlighted in red, green, blue, and cyan are also visible in the (a) left, (b) right, (c) back, and (d) top views, respectively.
(a): inverse-depth view from a dense \threed{} reconstruction; the regions visible in the left, and top views are respectively highlighted in green and cyan. 
(b): a view of the same scene as (a), 2\,m to the right;  the regions visible in the left, and top views are respectively highlighted in red and cyan. 
(c): a view of the same scene as (a), looking back; the region visible in top view is highlighted in cyan. 
(d): a view of the same scene as (a), looking down from 25\,m above; the regions visible in the left, right, and back views are respectively highlighted in red, green, and blue. 
Areas outside the highlighted regions are occluded – they are not visible from the other view.
For example, distant areas of (a)–(c), such as the sky, are not visible from the top view; the region behind the car on the right in (a) is not visible from (b).
These occluded parts of the image are ignored when computing the geometric consistency loss (Section~\ref{sec:method:model:loss}). 
For visualisation purposes, morphological closing has been applied to the visibility masks shown here, to remove some of the small-scale noise.}
\label{fig:viewpoints}
\label{fig:occlusion_mask}
\end{figure*}
\section{Related Work}\label{sec:related-work}
Several systems for building \threed{} reconstructions have been proposed, such as \borg{} \citep{Tanner:2015:BORG} or KinectFusion \citep{Newcombe:2011:KinectFusion}.
Our approach is to correct the \emph{output} of such a system by looking at the meshes it builds.
As we operate on inverse-depth images, our work is similar to depth refinement.
Below, we review some of the literature on that topic, as well as some other methods that our system draws inspiration from.


\subsubsection*{Learnt depth refinement and completion}
Some methods for \emph{learning} depth map refinement and completion have been recently proposed.
\cite{Uhrig:2017:SparsityInvariant,Eldesokey:2018:UncertaintySparsity} propose \ac{cnn} model for solving the KITTI Depth Completion Challenge, where sparse depth maps produced from laser data are densified.
\cite{Hua:2018:NConvDepthCompletion} use normalised convolutions \citep{Knutsson:1993:NormalizedConv} to predict dense depth maps that have been sparsely sampled.
These methods all rely on having some sort of high-quality, usually laser, depth information (albeit sparse) at run-time, and are only designed to fill in the missing data.
In contrast, we only require high-quality data during training, and our method aims to refine depth from a low-quality mesh in a more general sense – both filling in blanks, as well as refining existing surfaces.

A few other methods more closely related in spirit to ours aim to learn depth refinement using meshes as reference.
\cite{Kwon:2015:DataDrivenDepth} use dictionary learning to model the statistical relationships between raw \textsc{rgb-d} images, and high-quality depth data obtained by fusing multiple depth maps with KinectFusion \citep{Newcombe:2011:KinectFusion}.
Two other recent works use fused \textsc{rgb-d} reconstruction to obtain high-quality reference data and train \ac{cnn}s to enhance depth.
\cite{Zhang:2018:DeepDepthCompletion} uses a colour image to predict normals and occlusion boundaries, supervised by the \threed{} reconstruction, and then formulates an optimisation problem to fill in holes in an aligned depth image.
\cite{Jeon:2018:ReconstructionPairwiseDataset} introduce a 4000-image dataset of raw/clean depth image pairs, and train a \ac{cnn} to enhance raw depth maps.
Furthermore, they show that \threed{} reconstructions can be obtained with fewer data and quicker when using their depth-enhancing network.
These methods all rely on live colour images to guide the refinement or completion of live depth.
That means that only limited data is available for training.
In contrast, our purely mesh-based formulation allows us to extract many more training pairs from any viewpoint, removing any viewpoint-specific bias that might otherwise surface while learning.

\subsubsection*{Learnt depth estimation}
Another related line of research has been monocular depth estimation, which inspires our choices of network architecture and the use of geometry as a source of self-supervision.
\cite{Godard:2017:UnsupervisedMonoDepth,Laina:2016:Deeper,Ummenhofer:2017:DeMoN,Zhou:2017:SfMLearner,Dharmasiri:2018:ENG,Mahjourian:2018:Vid2Depth,Klodt:2018:SfMFromSfM}, etc. propose various \ac{cnn} models for depth estimation from single colour images.
When explicit depth ground-truth is unavailable, the training is self-supervised using multiview geometry: the predicted depths and relative poses should maximise the photometric consistency of nearby input images.
In this work, we show how the photometric consistency loss can be adapted to ensure consistency between inverse-depth predictions directly, without need for colour images.


\section{Method}\label{sec:method}

\subsection{Training data}\label{sec:method:training-data}
In contrast to many existing approaches, this work is about correcting \emph{existing} reconstructions, and therefore we assume the existence of \threed{} meshes of a scene.
In addition to the low-quality reconstruction we wish to correct, we also have a high-quality reconstruction of the same scene.
For example, we learn to correct \threed{} reconstructions from depth-maps using laser reconstructions as a reference.
To build the meshes we train on, we use the \borg{} \citep{Tanner:2015:BORG} system.

The training data consists of \twod{} views of the meshes. 
We use the same virtual camera to generate aligned views of the low-quality mesh and the high-quality mesh for each scene.
Because a mesh is available, we can generate multiple types of images at each viewpoint. 
In particular, we generate inverse-depth, normals, mesh triangle area, mesh triangle edge ratio (ratio between the shortest and the longest edge of each triangle), and surface-to-camera angle (Figure~\ref{fig:mesh_features}).
We refer to these images as \emph{mesh features}.
The ground-truth labels ($\gtruth$) are computed as the difference in inverse-depth between high-quality and low-quality reconstructions:

\begin{equation}\label{eq:g}
  \gtruth(p) = \hqidepth(p) - \lqidepth(p)
\end{equation}%
where $p$ is a pixel index, and $\hqidepth$ and $\lqidepth$ are inverse-depth images for the high-quality and low-quality reconstruction, respectively.
For notational compactness, $\gtruth(p)$ is referred to as $\gtruth$, and future definitions are over all values of $p$, unless otherwise mentioned. 

\ignore{An illustration of this is shown in Figure~.}

Inverse-depth is used instead of depth for several reasons. 
Firstly, it emphasises surfaces close to the camera where more information is available per pixel. 
Secondly, background (non-surface) pixels are not processed separately – they are assigned a value of zero, corresponding to points infinitely far away from the camera. 
If we used depth, those pixels would either have to be assigned an arbitrary finite value, which would result in semantic discontinuities in the output, or learnt to be ignored by the network, since there is no standard way to deal with infinite values in a \ac{cnn}.
Finally, another advantage of inverse-depth is that resampling it, which is needed to compute geometric consistency, is simpler than resampling depth images.

\subsection{Geometric consistency}\label{sec:method:geometric-consistency}
Intuitively, since the reconstructions we wish to correct are static, the predictions made from overlapping views should be geometrically consistent.
In other words, surfaces that appear in a certain location according to a prediction should appear in the same location in all predictions where they are in view.

We resample nearby predicted views according to the predicted geometry of the current view, and minimise the absolute difference in inverse-depth.
This dense warping is similar to reprojecting nearby views into the current view, but has the advantage of being differentiable, and of generating dense images instead of sparse reprojected pointclouds.
An illustration of this idea is shown in Figure~\ref{fig:geometric_consistency}.

Normally, dense warping is used in conjunction with colour images, where the values of the pixels are view-independent.
Since we are warping inverse-depth images, where the pixel values depend on the viewpoint, we need to compute the absolute difference in the same camera frame.

Concretely, for a view $t$, a nearby view $n$, let $\pred_t$ be a \ac{cnn} prediction for the target view, let $\pred_n$ be a prediction for the nearby view, and let $\predidepth_t = \lqidepth_t + \pred_t$ and $\predidepth_n = \lqidepth_n + \pred_n$ the corrected inverse-depth images for the two views. 
Furthermore, let $\mathbf{p}_t$ be pixel coordinates in the target view, $K$ the intrinsic matrix of the virtual camera, and $T_{n,t}$ the SE(3) transform from view $t$ to view $n$. 
We can then define the geometric inconsistency $\predidepth_{n,t} - \predidepthreproj_{n,t}$ where $\predidepth_{n,t}$ is the predicted inverse-depth from view $t$ in the frame of view $n$ and $\predidepthreproj_{n,t}$ is the warped inverse-depth from view $n$.

They are defined as follows:
\begin{align}
    \predidepthreproj_{n,t}(\mathbf{p}_t) & = \predidepth_n(\mathbf{p}_n), \\
    \predidepth_{n,t}(\mathbf{p}_t) & = \frac{\mathbf{x}_n^{(4)}}{\mathbf{x}_n^{(3)} + \epsilon},
    \label{eq:predidepthnt}
\end{align}
where $\mathbf{x}_n$ is the \threed{} homogeneous point in view $n$ corresponding to pixel $\mathbf{p}_t$, and $\mathbf{p}_n$ its projection:
\begin{align}
    \label{eq:pixn}
    \mathbf{p}_n &= \frac{1}{\mathbf{x}_n^{(3)} + \epsilon}\left(\mathbf{x}_n^{(1)}~\mathbf{x}_n^{(2)}\right)^T,\\
    \mathbf{x}_n &= F_h \mathbf{x}_t, \\
    \mathbf{x}_t &= \left(\mathbf{p}_t ~ 1 ~ \predidepth_t(\mathbf{p}_t)\right)^T, \\
    F_h &= K_h T_{n,t} K_h^{-1}, \\
    K_h &= \left(%
        \begin{array}{cc}%
            K & \mathbf{0} \\
            \mathbf{0}^T & 1
        \end{array}%
    \right).
\end{align}

The sample pixel $\mathbf{p}_n$ does not necessarily have integer coordinates, and thus may lie in-between pixels in the $\predidepth_n$ grid. 
Note that we add a small value $\epsilon$ with the same sign as $\mathbf{x}_n^{(3)}$ in Equations \ref{eq:predidepthnt} and \ref{eq:pixn} to avoid dividing by zero.
Under the mild assumption that surfaces between pixels are planar, we can sample $\predidepth_n$ by linearly interpolating the four pixels nearest to $\mathbf{p}_n$ – another advantage of the inverse-depth formulation.

\subsubsection*{Occlusion masks}
A common problem with this approach is that views cannot be consistent in the presence of occlusions.
In our setting, however, since the views are synthetic (and therefore the intrinsic and extrinsic parameters are perfectly known), we are able to create occlusion masks and only apply the geometric consistency loss where there are no occlusions.

We compute occlusion masks from the reference high-quality reconstruction.
Each mesh triangle is assigned an index by hashing its world frame coordinates.
Then, in addition to mesh features, an image with mesh triangle indices is generated at each location.
For each pair of views, each pixel of the triangle index image is resampled from one view into the other in a similar fashion to the inverse-depth above.
Instead of interpolating between the four nearest neighbours, these are returned as four separate samples.
The pixels where the indices match at least one of the four samples are considered unoccluded, and the pixels where they do not are considered occluded (see Figure~\ref{fig:occlusion_mask}).

Small errors appear due to rasterisation when mesh triangles are too small, often those that are far away from the camera.
We could avoid these errors by generating occlusion masks with the OpenGL rasterisation pipeline when the rest of the data is generated.
However, this greatly increases the space required to store training data – the number of occlusion masks scales quadratically with the number of views we want to enforce consistency between.
Generating the occlusion masks on the fly means that geometric consistency can be enforced between arbitrary views, so more settings can be explored without regenerating part of the training data.
In Section~\ref{sec:experiments} we show that the quality of the occlusion masks is sufficient to demonstrate the advantages of geometric consistency.

\subsection{Model}\label{sec:method:model}

\subsubsection{Network architecture}\label{sec:method:model:network-architecture}
The model used is an encoder-decoder \ac{cnn} similar to the one proposed in \cite{Tanner:2018:MeshedUp}. 
The encoder is composed of residual blocks based on the ResNet-50 architecture \citep{He:2016:ResNet}, and  the decoder uses up-convolutions proposed by \cite{Shi:2016:PixelShuffle}. 
U-Net \cite{Ronneberger:2015:UNet} style skip connections are added between the encoder and the decoder to improve the sharpness of predictions. 
As a simple way to offer our model some introspective capabilities, we predict a soft attention mask (with values $\in [0, 1]$) in addition to the error in inverse-depth.
This mask is multiplied pixel-wise with the error prediction to modulate which parts are going to be used and does not require extra supervision.
Table~\ref{tab:network-architecture} provides an overview for each of the layers of the proposed \ac{cnn}.

\begin{table}[t!]
    \centering
    \caption{Overview of the cnn architecture for error prediction}
    \begin{tabular}{p{0.40\linewidth}cc}
        \toprule
        \textbf{Block Type} & \textbf{Filter Size/Stride} & \textbf{Output Size} \\
        \midrule
        Input & - & $96\times 288\times F$ \\
        Convolution, Residual & $3\times 3$/$1$ & $96\times 288\times 64$ \\
        Convolution & $5\times 5$/$2$ & $48\times 144\times 64$ \\
        Max Pool & $3\times 3$/$2$ & $24\times 72\times 64$ \\
        Residual$\times2$, Projection & $3\times 3$/$2$ & $12\times 36\times 256$ \\
        Residual$\times2$, Projection & $3\times 3$/$2$ & $6\times 18\times 512$ \\
        Residual$\times2$, Projection & $3\times 3$/$2$ & $3\times 9\times 1024$ \\
        Residual$\times8$ & $3\times 3$/$1$ & $3\times 9\times 2048$ \\
        Up-projection & $3\times 3$/$\frac{1}{2}$ & $6\times 18\times 1024$ \\
        Up-projection & $3\times 3$/$\frac{1}{2}$ & $12\times 36\times 512$ \\
        Up-projection & $3\times 3$/$\frac{1}{2}$ & $24\times 72\times 256$ \\
        Up-projection & $3\times 3$/$\frac{1}{2}$ & $48\times 144\times 128$ \\
        Up-projection & $3\times 3$/$\frac{1}{2}$ & $96\times 288\times 32$ \\
        Residual & $3\times 3$/$1$ & $96\times 288\times 32$ \\
        Convolution & $3\times 3$/$1$ & $96\times 288\times 2$ \\
        \bottomrule
    \end{tabular}
\label{tab:network-architecture}
\end{table}

\subsubsection{Loss}\label{sec:method:model:loss}
The objective function has several components, as follows. 
The first term is the data loss that minimises the error between the output and the label.
To compute this loss, we use berHu norm \citep{Owen:2007:BerHu}.
For large errors, this behaves in the same way as an $L_2$ norm.
For small errors, where the gradients of $L_2$ become too small to drive the error completely to zero, $L_1$ norm is used instead.
The advantages of this norm have also been observed in \cite{Laina:2016:Deeper,Ma:2018:SparseToDense}. 
The data loss is defined as follows:

\begin{equation}
  \loss{\dataname} = %
    \sum_{p \in V} \weight \cdot \left\|\pred - \gtruth\right\|_{berHu}, 
\end{equation}%
where $p$ is the pixel index, $V$ is the set of valid pixels (to account for missing data in the ground-truth), $\weight$ is a per-pixel weight detailed in Section~\ref{sec:method:model:loss-weight}, $\pred$ and $\gtruth$ are the prediction and the target, respectively, and $\|\cdot\|_{berHu}$ is the berHu norm. 


To improve small-scale details and prevent artefacts in the prediction, while also allowing for sharp discontinuities, we also apply a loss on the gradient of the predictions:
\begin{equation}
   \loss{\gradname} = \frac{1}{2} \sum_{p \in V} \weight \cdot %
   \left(\left|\partial_x \pred{} - \partial_x \gtruth{} \right| %
    + \left|\partial_y \pred{} - \partial_y \gtruth{} \right| \right). %
\end{equation}%

We use the Sobel operator \citep{Sobel:1968} to approximate the gradients in the equation above.

The geometric consistency loss guides nearby predictions to have the same \threed{} geometry, and relies on reprojected nearby views $\predidepthreproj_{n,t}$. 
For a target view $t$, a set of nearby views $N$, the set of pixels unoccluded in a nearby view $U_n$ (see Figure~\ref{fig:occlusion_mask}), this loss is defined as:
\begin{equation}
   \loss{\gcname} = \sum_{n \in N}\sum_{p \in U_n} \left|\predidepthreproj_{n,t} - \predidepth_{n,t}\right|.
\end{equation}
Note that $U_n$ has no relation to the set of valid pixels ($V$) from the previous losses, since this loss is only computed between predictions.
This enables the network to make sensible predictions even in parts of the image which have no valid label.

Finally, we also include an $L_2$ weight regulariser, $\loss{\regname}$, to reduce overfitting by keeping the weights small. The overall objective is thus defined as:
\begin{equation}
    \loss{} = %
        \scaling{\dataname} \loss{\dataname}_s %
        + \scaling{\gradname} \loss{\gradname} %
        + \scaling{\gcname} \loss{\gcname} %
        + \scaling{\regname} \loss{\regname}, %
\end{equation}
where $s$ is the scale, and the $\lambda$s are weights for each of the components (see Table~\ref{tab:parameters} for values).
\begin{table}
\caption{Summary of Hyperparameters Used in System}
\label{tab:parameters}
\begin{tabular}{llp{5cm}}
\toprule
\textbf{Symbol}         & \textbf{Value}                & \textbf{Description} \\
\midrule
$\scaling{\dataname}$   & 1                   & Weight of the data loss. \\
$\scaling{\gradname}$   & 0.1                 & Weight of the smoothness loss. \\
$\scaling{\gcname}$     & 0.1                 & Weight of the geometric consistency loss. \\
$\scaling{\regname}$    & $10^{-6}$           & Weight of the $L_2$ variable regulariser. \\
$w_{min}$                 & 0.1               & Minimum per-pixel loss scaling. \\
$w_{max}$                 & 5                 & Maximum per-pixel loss scaling. \\
\midrule
$\eta_{max}$ & $10^{-4}$ & Initial learning rate. \\
$\eta_{min}$ & $5\cdot10^{-6}$ & Final learning rate. \\
$T_{max}$ & $1.2\cdot10^5$ & Learning rate decay steps. \\
$\beta_1$ & 0.9 & Adam exponential decay rate for first moment estimates. \\
$\beta_2$ & 0.999 & Adam exponential decay rate for second moment estimates. \\
 & 80 & Norm at witch gradients are clipped during training. \\
 & 16 & Batch size. \\
 & $5\cdot10^5$ & Number of training steps. \\
\bottomrule
\end{tabular}
\end{table}

\subsubsection{Loss weight}\label{sec:method:model:loss-weight}
Inspired by the work of \cite{Ronneberger:2015:UNet} on U-Nets, we use a loss-weighting mechanism based on the Euclidean Distance Transform \citep{Felzenszwalb:2012:DistanceTransforms} to emphasise edge pixels when regressing to the error in depth.
We first extract Canny edges \citep{Canny:1986:EdgeDetect} from the ground-truth labels. Based on these edges, we then compute the per-pixel weights as:
\begin{equation}
    \begin{array}{l}
         d(p) = \mathrm{ln}(1 + \mathrm{EDT}(p)) \\
         \weight{}(p) = (w_{max} - w_{min}) \left(1 - \frac{d(p)}{\max_{p} d(p)}\right) + w_{min}
    \end{array},
\end{equation}%
where $\weight{}(p)$ is the loss weight for pixel $p$, $\mathrm{EDT}(p)$ is the Euclidean Distance Transform at pixel $p$, and $w_{min}$ and $w_{max}$ are the desired range of the per-pixel weight.

\section{Experiments}\label{sec:experiments}
\begin{figure}[t!]
\centering
\includegraphics{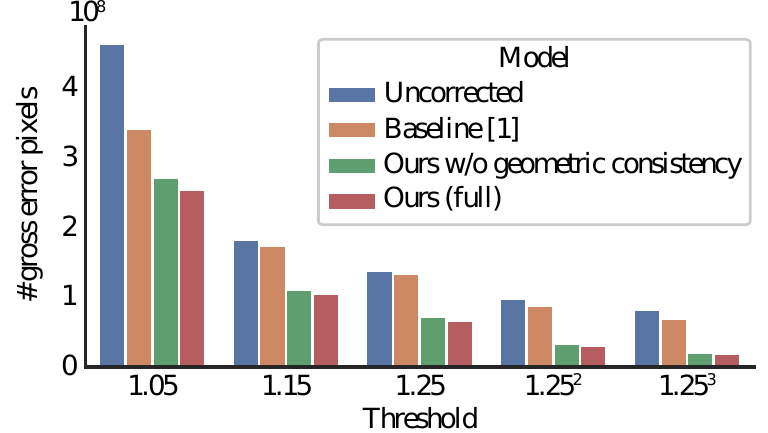}
\caption{%
Gross error correction.
For each threshold (Equation~\ref{eq:thresholded_accuracy}), we count how many pixels are incorrect in the predictions over the test set.
Our full model removes 45.3\% of the smaller errors and 77.5\% of the gross errors.
The baseline model is unable to effectively handle the multi-view setup, and fails to correct gross errors.
}
\label{fig:gross_errors}
\end{figure}

\subsection{Experimantal setup}
\begin{table*}[h]
\centering
\caption{Generalisation Capability of Depth Error Correction}
\label{tab:results_generalisation}
\begin{filecontents*}{tables/results-generalisation.csv}
Model, Train, Test, imae, irmse, thacc1.05, thacc1.15, delta1, delta2, delta3 
Uncorrected,        –, 00, 0.017593,0.081029,0.527110,0.789971,0.840150,0.887261,0.907113
 \dittotikz,        –, 05, 0.015702,0.081510,0.551061,0.806109,0.842725,0.882996,0.904586
 \dittotikz,        –, 06, 0.028137,0.124019,0.494694,0.727073,0.750705,0.786409,0.804902
       Ours, {05; 06}, 00, 0.014467,0.080329,0.661392,0.852655,0.899259,0.947568,0.968180
 \dittotikz, {00; 06}, 05, 0.011927,0.076691,0.703784,0.863107,0.902484,0.948622,0.969933
 \dittotikz, {00; 05}, 06, 0.023694,0.115290,0.622739,0.781407,0.817207,0.858488,0.881995
\end{filecontents*}
\pgfplotstabletypeset[
    columns/Test/.style={ column type={c|}, },
    every row no 2/.style={after row=\midrule}
]{tables/results-generalisation.csv}
\end{table*}

\subsubsection*{Training and inference}\label{sec:experiments:training-and-inference}
The network is implemented in Python using TensorFlow~v1.12.
Each model is trained on an Nvidia Titan V GPU.
The weights are optimised for 500\,000 steps with a batch size of 16 using the Adam \citep{Kingma:2015:Adam} optimiser.
The learning rate ($\eta_t$) is decayed linearly for the first 120\,000 steps.
Generating the mesh features takes an average of 52\,ms per view using OpenGL on an Nvidia GTX Titan Black, and inference takes an average of 12.5\,ms on the Titan V.
All the training hyper-parameters are defined in Table~\ref{tab:parameters}.

\subsubsection*{Dataset}
Three sequences from the \ac{kitti-vo} dataset were used as the input to the reconstruction pipeline.
For each sequence, two reconstructions are built: one from the stereo camera depth-maps, and one from the laser data.
Both the meshes are generated with a fixed voxel width of 0.2\,m.
In the experiments, we show how to learn a correction of the depth-map reconstruction using the laser reconstruction as reference.
Using \ac{glsl}, we create a virtual camera and project each dense reconstruction into mesh features.
We sample locations along the original trajectory in each sequence every 0.3m, and at each location we generate mesh features from four different viewpoints.
An illustration of the viewpoints is shown in Figure~\ref{fig:viewpoints}.
For all experiments, the \ac{kitti-vo} sequences 00, 05, and 06 are used, from which a total of 96\,728 training examples of size $96\times288$ are generated.
We split each of the mesh feature sequences into three distinct parts: the first 80\% we use as training data, the next 10\% we use for validating hyperparameter choices, and the last 10\% we use for evaluation.
All three sequences are predominantly in urban environments with small amounts of visible vegetation.

\subsubsection*{Performance metrics}
We use some metrics common in literature for assessing inverse-depth predictions.

Our first metric measures the accuracy of our network’s ability to estimate errors under a given threshold, serving as an indication of how often our estimate is correct.
The thresholded accuracy measure is essentially the expectation that a given pixel in $V$ is within a threshold $thr$ of the label:
\begin{equation}\label{eq:thresholded_accuracy}
    \delta = \mathbb{E}_{p \in V} \left[
        \mathbb{I}\left(\textrm{max}\left(
            \frac{\hqidepth}{\predidepth}, 
            \frac{\predidepth}{\hqidepth}\right) 
        < thr\right)
    \right],
\end{equation}
where $\hqidepth$ is the reference inverse-depth, $\predidepth$ is the predicted inverse depth, $V$ is the set of valid pixels, and $n$ is the cardinality of $V$, and $\mathbb{I(\cdot)}$ represents the indicator function. For granularity, we use $thr = \in \{1.05, 1.15, 1.25, 1.25^2, 1.25^3\}$. 

In addition, the \ac{mae} and \ac{rmse} metrics provide a quantitative measure of per pixel error and are computed as follows:
\begin{align}
\textrm{iMAE} &= \frac{1}{n}\sum_{p \in V}{\left|\predidepth - \hqidepth\right|}, \\
\textrm{iRMSE} &= \sqrt{\frac{1}{n}\sum_{p \in V}{(\predidepth - \hqidepth)^2}},
\end{align}
where the ‘i’ indicates that the metrics are computed over inverse-depth images.

\subsection{Gross error correction}\label{sec:experiments:gross-error}
We first look at how well our model corrects gross errors in inverse-depth.
Equation~\ref{eq:thresholded_accuracy} can be used to classify pixels in an image as either correct or incorrect at a given threshold.
Using this method, we count the number of incorrect pixels in our predictions, and compare it to the number of incorrect pixels in the input inverse-depth.
As a baseline, we train the model proposed in \cite{Tanner:2018:MeshedUp} on our dataset, and compare it to our full model, trained with and without the geometric consistency loss (Figure~\ref{fig:gross_errors}).
Our proposed model outperforms the baseline at correcting both small errors ($thr = 1.05$) as well as larger errors, reducing the number of errors at $thr = 1.25^3$ by 77.5\%.

\subsection{Generalisation capability}
To be useful in practice, the model needs to be able to generalise to new data. 
For example, separate models could be trained for indoor scenes and outdoor scenes, or other different types of environments, but certainly within the same kind of environment, one model should work well for a variety of scenes. 

We evaluate the ability of our proposed model to generalise by training the full model on a subset of the available sequences, and testing it on the rest (excluding the frames used for validation in Section~\ref{sec:experiments:gross-error}).
Table~\ref{tab:results_generalisation} shows the data splits and the performance of the models on the test sequences.
Our model performs particularly well on sequence 06, where the input reconstruction has very large errors. This highlights the ability of our model to reduce gross errors.

\subsection{Ablation study}\label{sec:experiments:component-analysis}
\begin{table*}
\centering
\caption{Depth Error Correction Ablation Study Results}
\label{tab:results_progressive}
\begin{filecontents*}{tables/results-progressive.csv}
Model                                     ,imae,irmse,thacc1.05,thacc1.15,thacc1.25,thacc1.25^2,thacc1.25^3
Uncorrected                               ,0.01852627471089363,0.0872877687215805,0.5090488791465759,0.807686984539032,0.8537503480911255,0.8972411751747131,0.9134893417358398
Baseline                                  ,0.014741498790681363,0.06254323422908784,0.6392216682434082,0.8165080666542053,0.8589719891548157,0.9072268724441528,0.9269263744354248
\hspace{2mm}w/ our losses; no \textsc{gc} ,0.012828195840120316,0.06262855231761932,0.6776933193206787,0.843455147743225,0.8905291080474853,0.9415768384933472,0.9616763353347778
\hspace{2mm}w/ our losses; \textsc{gc}    ,0.011668421886861324,0.05435284748673439,0.6865504503250122,0.8463888168334961,0.8945192933082581,0.948083472251892,0.9665467500686645
Ours (no attn; no \textsc{gc})            ,0.011199291609227658,0.05486475601792336,0.7046730399131775,0.8768023490905762,0.9208074808120728,0.9634630084037781,0.9771822690963745
Ours (no \textsc{gc})                     ,0.011219736002385617,0.05626401379704475,0.7131281614303588,0.8833992958068848,0.9241024255752563,0.9640994071960449,0.9780895113945007
Ours (no attn)                            ,0.010201134532690049,0.053129959106445315,0.7293243885040284,0.8927148103713989,0.9320035576820374,0.968650221824646,0.9810054779052735
Ours (full)                               ,0.010560845583677292,0.05365167111158371,0.731573760509491,0.8894312977790833,0.9291235923767089,0.9676058053970337,0.9805144429206848
\end{filecontents*}
\pgfplotstabletypeset[
    string replace={0}{},
    columns/Model/.style={ column type={l|}, }, 
    every row no 3/.style={ after row=\midrule, },
]{tables/results-progressive.csv}
\end{table*}
\newcommand{\gqual}[1]{%
\includegraphics[width=0.16\textwidth,frame]{figures/graphics/qualitative/#1.png}}
\newcommand{\blnk}{%
\includegraphics[width=0.16\textwidth,cframe=white]{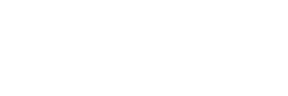}}
\newcommand{\perr}[3]{\gqual{#1/#2-predicted_error_#3}}
\newcommand{\pdepth}[3]{\gqual{#1/#2-predicted_idepth_#3}}
\newcommand{\pmask}[3]{\gqual{#1/#2-softmask_#3}}
\newcommand{\labs}[2]{
\gqual{labels/#1-lq_inverse_depth_#2}\\[0.6ex]
\gqual{labels/#1-masked_gterror_#2}\\[0.6ex]
\gqual{labels/#1-hq_inverse_depth_#2}
}
\newcommand{\preds}[3]{
\blnk\\[0.6ex]
\perr{#1}{#2}{#3}\\[0.6ex]
\pdepth{#1}{#2}{#3}
}
\newcommand{\predsmask}[3]{
\pmask{#1}{#2}{#3}\\[0.6ex]
\perr{#1}{#2}{#3}\\[0.6ex]
\pdepth{#1}{#2}{#3}
}
%
\begin{figure*}[t]
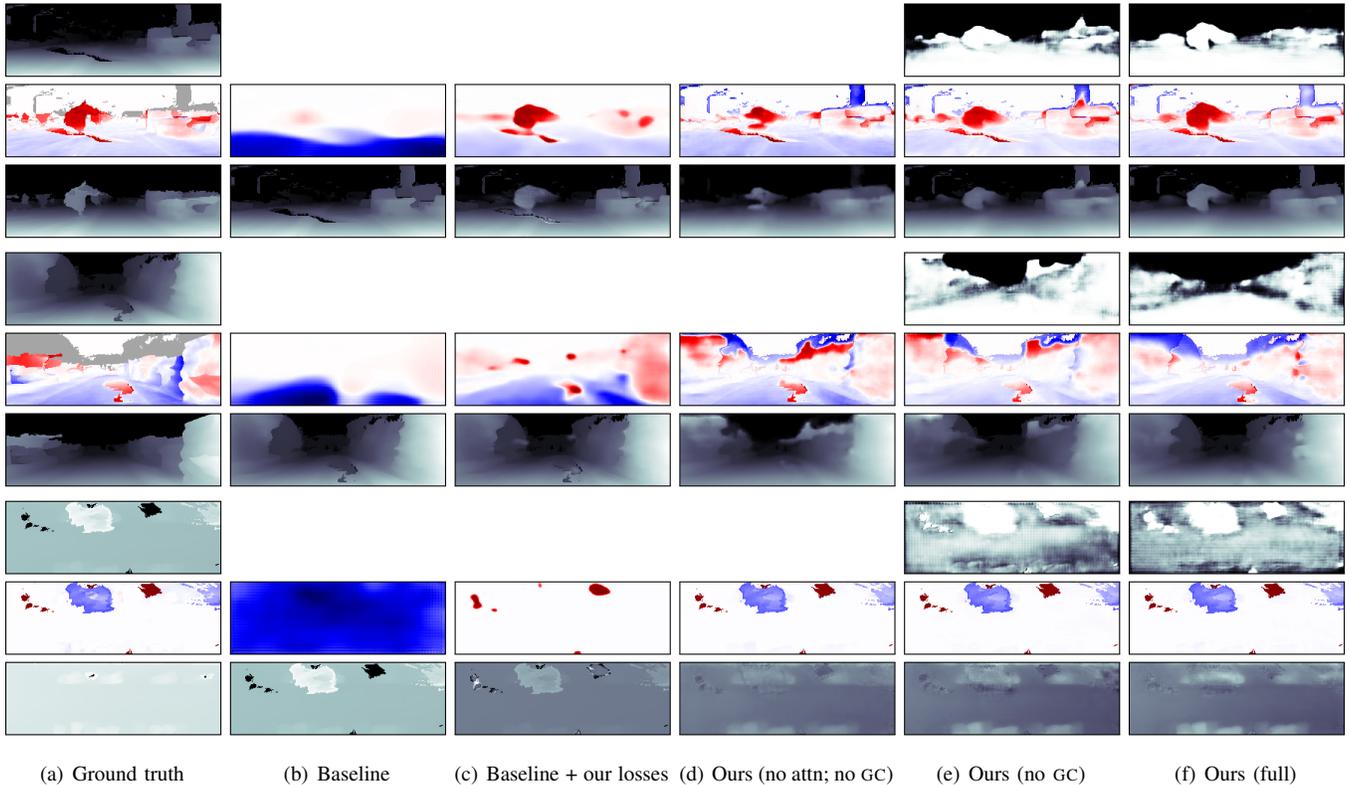

\centering
\subfigure[Ground truth]{%
\parbox{0.16\textwidth}{
\labs{019}{0}\\[1.2ex]
\labs{004}{0}\\[1.2ex]
\labs{118}{3}
}}\hfill
\subfigure[Baseline]{%
\parbox{0.16\textwidth}{
\preds{v0-baseline}{019}{0}\\[1.2ex]
\preds{v0-baseline}{004}{0}\\[1.2ex]
\preds{v0-baseline}{118}{3}
}}\hfill
\subfigure[Baseline + our losses]{%
\parbox{0.16\textwidth}{
\preds{v0-finetuned+gcexp}{019}{0}\\[1.2ex]
\preds{v0-finetuned+gcexp}{004}{0}\\[1.2ex]
\preds{v0-finetuned+gcexp}{118}{3}
}}\hfill
\subfigure[Ours (no attn; no \textsc{gc})]{%
\parbox{0.16\textwidth}{
\preds{v1-full}{019}{0}\\[1.2ex]
\preds{v1-full}{004}{0}\\[1.2ex]
\preds{v1-full}{118}{3}
}}\hfill
\subfigure[Ours (no \textsc{gc})]{%
\parbox{0.16\textwidth}{
\predsmask{v1-full_softmask}{019}{0}\\[1.2ex]
\predsmask{v1-full_softmask}{004}{0}\\[1.2ex]
\predsmask{v1-full_softmask}{118}{3}
}}\hfill
\subfigure[Ours (full)]{%
\parbox{0.16\textwidth}{
\predsmask{v1-full_softmask+gcexp}{019}{0}\\[1.2ex]
\predsmask{v1-full_softmask+gcexp}{004}{0}\\[1.2ex]
\predsmask{v1-full_softmask+gcexp}{118}{3}
}}
\caption{%
Illustration of how the quality of the predictions changes as different components are added to the system. Each three rows show an example, as follows:
(a) Input inverse-depth and ground-truth error, and ground-truth inverse-depth. The shaded areas in the ground-truth error represent missing data in the reference mesh. (b)–(d) Prediction and corrected inverse-depth. (e), (f): Attention mask, prediction, and corrected inverse-depth.
The baseline model (b) only learns very rough predictions and is unable to generalise well to the top viewpoint (last example).
Our proposed losses help with generalisation across viewpoints, but without skip connections in the network predictions are not very well localised (c).
Our model (d)–(f) makes well-localised predictions.
The attention mask removes some of the spurious predictions where there is no reference data (e), and geometric consistency further guides this (f).
}
\label{fig:results_qualitative}
\end{figure*}
To better understand how different components of our model improve the learnt correction, we perform an ablation study.
The results in Table~\ref{tab:results_progressive} show that our proposed geometric consistency loss (rows with \textsc{gc}) improves performance at all error scales.

The proposed attention mask (rows with attn) improves the performance in the absence of geometric consistency, but slightly limits the performance especially with larger errors.
However, qualitatively (Figure~\ref{fig:results_qualitative}), the attention mask allows us to better handle missing training data. In particular, surfaces are not spuriously removed or added in those regions: the model learns to mask those regions out of the correction and keep them as they are.

\section{Conclusion}
In this paper we present a method for correcting gross errors in dense \threed{} meshes.
We extracted paired \twod{} mesh features from two reconstructions and trained a neural network to predict the difference in inverse-depth between the two.
We addressed the issue of overly-smooth predictions with a U-Net architecture and a loss-weighting mechanism that emphasises edges.
The geometric consistency of our predictions is improved with a view-synthesis loss that targets inconsistencies.
Our experiments show that the proposed method reduces gross errors in inverse-depth views of the mesh by up to 77.5\%.

\section*{Acknowledgment}
The authors would like to acknowledge the support of the UK’s Engineering and Physical Sciences Research Council (\textsc{epsrc}) through the Centre for Doctoral Training in Autonomous Intelligent Machines and Systems (\textsc{aims}) Programme Grant EP/L015897/1.
Paul Newman is supported by EPSRC Programme Grant EP/M019918/1.
The authors would like to acknowledge the use of the University of Oxford Advanced Research Computing (ARC) facility\footnote{\url{http://dx.doi.org/10.5281/zenodo.22558}} in carrying out this work.

\bibliographystyle{IEEEtran}
\bibliography{references}
~ 

\end{document}